\pdfoutput=1
\documentclass[11pt]{article}

\usepackage{acl}

\usepackage{times}
\usepackage{latexsym}

\usepackage{graphicx}
\usepackage{amsmath}
\usepackage{amsfonts}
\usepackage{multirow}
\usepackage{amsfonts}

\usepackage{siunitx}
\usepackage{colortbl}
\usepackage{makecell}
\usepackage[ruled,linesnumbered]{algorithm2e}
\usepackage{algpseudocode} 
\usepackage{subcaption}
\usepackage{bbm}
\definecolor{maroon}{cmyk}{0,0.87,0.68,0.32}

\usepackage[T1]{fontenc}
\usepackage[utf8]{inputenc}

\usepackage{microtype}

\title{UNIMO-2: End-to-End Unified Vision-Language Grounded Learning}

\author{Wei Li, Can Gao, Guocheng Niu, Xinyan Xiao, \\ \textbf{Hao Liu, Jiachen Liu, Hua Wu, Haifeng Wang} \\
  Baidu Inc., Beijing, China \\
  \texttt{\{liwei85,gaocan01,niuguocheng,xiaoxinyan,} \\
  \texttt{liuhao24,liujiachen,wu\_hua,wanghaifeng\}@baidu.com}
  }

\begin{document}
\maketitle
\begin{abstract}

Vision-Language Pre-training (VLP) has achieved impressive performance on various cross-modal downstream tasks.
However, most existing methods can only learn from aligned image-caption data and rely heavily on expensive regional features, which greatly limits their scalability and performance.
In this paper, we propose an end-to-end unified-modal pre-training framework, namely UNIMO-2, for joint learning on both aligned image-caption data and unaligned image-only and text-only corpus.
We build a unified Transformer model to jointly learn visual representations, textual representations and semantic alignment between images and texts.
In particular, we propose to conduct grounded learning on both images and texts via a sharing grounded space, which helps bridge unaligned images and texts, and align the visual and textual semantic spaces on different types of corpora.
The experiments show that our grounded learning method can improve textual and visual semantic alignment for improving performance on various cross-modal tasks.
Moreover, benefiting from effective joint modeling of different types of corpora, our model also achieves impressive performance on single-modal visual and textual tasks. Our code and models are public at the UNIMO project page \url{https://unimo-ptm.github.io/}.
%not only achieves much better performance on joint vision-language tasks, but

%Especially, the zero-shot image/text retrieval performance achieves absolute gains of 10\% R@1 score on Flickr-30k test split.
%However, existing pre-training methods only learn from annotated image-text pairs but cannot effectively utilize large-scale non-paired texts and images, which greatly restrict their performance upper limit and modality adaptability.

%However, existing pre-training methods suffer from several main drawbacks: (1) only learn from annotated image-text pairs but cannot effectively utilize large-scale non-paired texts and images, (2) align textual and visual semantic space by coarse-level image-text matching without capturing fine-grained granularity of semantics across different modalities, (3) heavily rely on image feature extraction processes, such as region features by specific object detector.

\end{abstract}

\section{Introduction}

Large-scale pre-training has drawn much attention in the community of Computer Vision (CV), Natural Language Processing (NLP) and Multi-Modal (MM) due to its strong capability of generalization and efficient usage of large-scale data.
However, in the existing literature, the work on vision, language and vision-language representation learning are mostly studied separately with different training data sources.
In the vision domain, pre-training on large-scale image corpus such as ImageNet \citep{deng2009imagenet}, OpenImages \citep{kuznetsova2020open} and JFT-300M \citep{dosovitskiy2020image} has proven to be critical for learning transferring visual representation for various downstream tasks.
In NLP, pre-training on easily-accessible unannotated text corpora greatly improves the capabilities of language understanding and generation \citep{devlin-etal-2019-bert, liu2019roberta, yang2019xlnet}.
Pre-training has also become the de-facto approach in vision-language modeling \citep{lu2019vilbert, chen2020uniter, li2020oscar, li2019unicoder, yu2020ernie}. However, existing VLP methods require a massive amount of aligned image-text pairs which are costly to collect and hard to scale up. %The scales of these datasets are limited, which is at least an order of magnitude smaller than the vision domain, and much smaller than the large text corpora for NLP pre-training.
The large volumes of image corpus in CV and text corpus in NLP cannot be effectively utilized.
Thus, the scalability and performance upper limit of existing VLP methods are largely restricted.
As they only learn joint vision-language representations on image-text pairs, they are also difficult to be effectively adapted to visual and textual tasks \citep{li2020unimo, lin2020interbert}.
%Moreover, most of them rely heavily on expensive region-based visual feature extraction by pre-trained object detector, so they face the problems of limited visual expressive power and computation inefficiency.

%Most VLP methods only focus on joint language-vision tasks, but is difficult to be effectively adapted to visual and textual tasks (Li et al., 2020; Lin et al, 2020). Learning from large volumes of non-aligned images and texts is promising to greatly boost their performance and scalability on different types of tasks. Furthermore, most existing mainstream VLP models rely heavily on expensive image feature extraction processes, which firstly extract semantic visual features using a pre-trained object detection model, and then utilize the derived object-centric representation of the image for cross-modal pre-training.
%Furthermore, most existing VLP methods rely heavily on the expensive image feature extraction processes, most of which employ an object detector pre-trained on the Visual Genome dataset (Krishna et al., 2017) annotated with 1,600 object classes and 400 attribute classes as in Anderson et al. (2018).
%The object detection model is usually trained on specific visual dataset such as Visual Genome dataset (Krishna et al., 2017) annotated with 1,600 object classes and 400 attribute classes (Anderson et al., 2018).
%The processes of extracting regional features are also very time-consuming.
%Thus, they also face the problems of limited visual expressive power and computation inefficiency.

To address the limitations, we propose a new end-to-end unified-modal pre-training framework, namely UNIMO-2, for joint learning on various types of corpora, including images, texts, and image-caption pairs.
Specifically, we build a unified Transformer model to jointly learn visual representations, textual representations, and cross-modal alignment from the three types of corpora.
Both the visual and textual representations are learned end-to-end from raw images and textual sentences.
%Most existing VLP methods only learn joint vision-language representations on image-text pairs, so they are difficult to be effectively adapted to visual and textual tasks \citep{li2020unimo, lin2020interbert}.
Combining a large number of unaligned images and texts is not only expected to improve the performance of joint vision-language tasks, but also improve the scalability of adapting to single-modal visual and textual tasks.
%However, it is challenging to end-to-end bridge the unaligned visual and text corpora so as to effectively align the visual and text semantic space on different types of corpus.
However, it is challenging to bridge unaligned images and texts and effectively align the visual and textual semantic spaces on different types of corpora.

Only a few works have attempted to bridge unaligned images and texts by leveraging object tags from an pre-trained object detector as ``anchor points'' \citep{li2020unsupervised, li2020unimo}.
However, they all rely heavily on expensive object-centric visual feature extraction, thus facing the problems of limited visual expressive power and computation inefficiency.
%\citet{li2020unsupervised} has explored unsupervised VLP with unaligned image and text corpora by leveraging the object tags from an object detector as ``anchor points''. %They first extract region-based visual features and append the detected object tags to the visual input to encourage cross-modal fusion.
%UNIMO \citep{li2020unimo} propose to model images, text and image-text pairs jointly based on regional object extraction and prediction.
%cross-modal contrastive learning with complex data argumentation techniques.
%, and align visual and textual semantic spaces by cross-modal contrastive learning with complex data argumentation techniques.
%They both heavily rely on pre-extraction of region-based visual features and object tags.
%However, the visual feature extraction process is very time-consuming, and the predefined object detector limits the visual expressive power. 
%The object detection model is usually trained on specific visual datasets such as the Visual Genome dataset (Krishna et al., 2017), which annotates 1,600 object classes and 400 attribute classes (Anderson et al., 2018).
In this paper, in order to bridge the unpaired image and text corpora and align the visual and textual semantic spaces end-to-end, we propose to conduct grounded learning on images, texts, and image-text pairs via a sharing grounded space.
Specifically, we introduce a grounded dictionary shared by images and texts, which represents vision-language grounded semantics.
To learn the grounded dictionary, we apply vector quantization on both visual and textual representations to group image patches and text tokens with similar semantics into grounded tokens.
Furthermore, we design a Grounded Transformer architecture to let the visual and textual information exchanged by the grounded tokens, which not only facilitates grounded dictionary learning, but also improves cross-modal alignment.
Our grounded learning method can help bridge the textual and visual semantic spaces on unpaired image and text corpora to improve cross-modal fusion on different types of corpora.

We evaluate UNIMO-2 on a variety of representative vision-language understanding and generation tasks, including image/text retrieval, visual question answering, visual reasoning and image caption. On all these tasks, UNIMO-2 obtains obvious improvements compared to the baselines that only learn on aligned image-caption data or without our grounded learning component.
Moreover, we also evaluate our model on single-modal textual tasks such as natural language inference and visual tasks such as image classification \citep{deng2009imagenet}. %Specifically, the textual tasks include sentiment classification on the SST-2 dataset (Socher et al., 2013), natural language inference on the MNLI dataset (Williams et al., 2017), linguistic acceptability analysis on the CoLA dataset (Warstadt et al., 2019) and semantic similarity analysis on the STS-B dataset (Cer et al., 2017). The visual tasks include image classification on ImageNet.
The results show that our model has also achieved very impressive performance on these tasks, which proves the strong scalability and adaptability of our model.

UNIMO-2 has the following advantages compared with previous methods:
\begin{itemize}
    \item UNIMO-2 can jointly learn from both aligned and unaligned image and text corpora end-to-end, effectively alleviating the limitations of corpus, and learning more generalized visual and textual representations on large volumes of different types of corpus.
    \item Benefiting from utilizing different types of corpora, UNIMO-2 has better scalability for different types of tasks, including both cross-modal tasks and single-modal tasks.
    \item Our grounded learning method can help align textual and visual semantic spaces more effectively, thereby greatly improving the performance of various cross-modal tasks. In particular, the performance of zero-shot image/text retrieval even outperforms CLIP pre-trained on an order of magnitude larger pair corpus.
\end{itemize}

\begin{figure*}[ht!]
	\centering
	\includegraphics[width=6.3in]{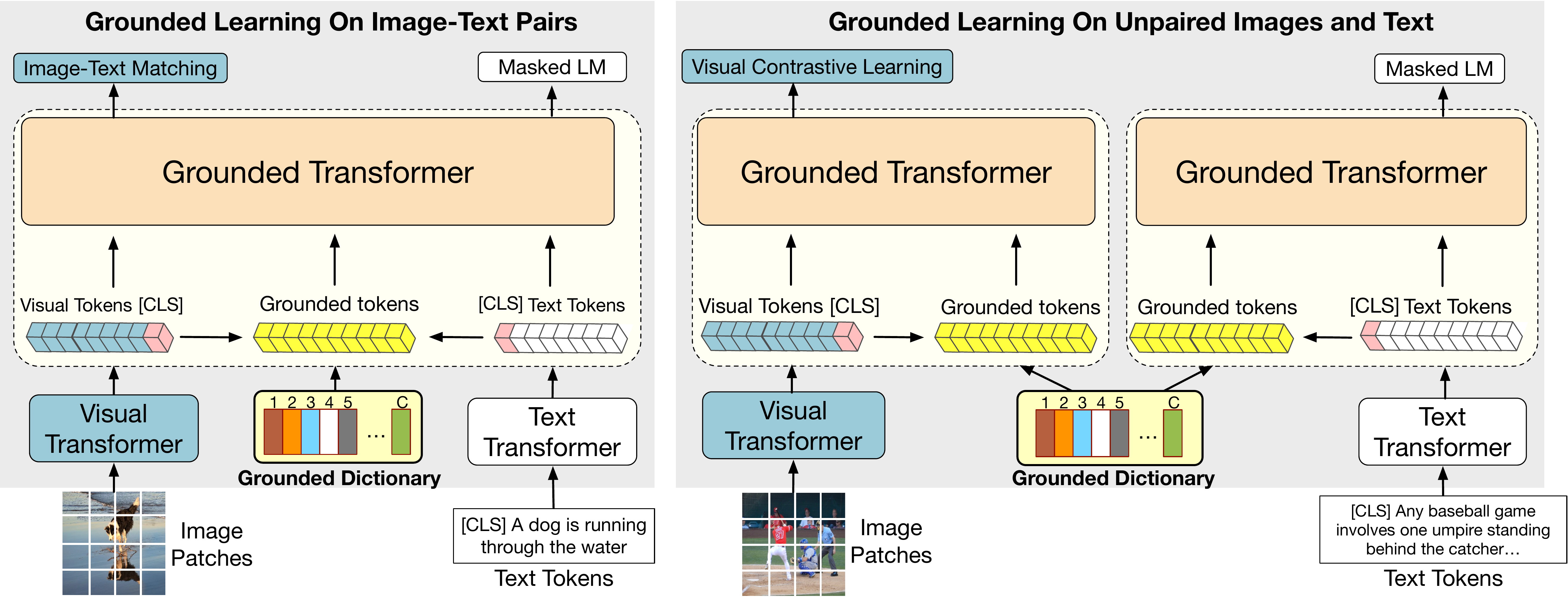}
	\caption{Illustration of our UNIMO-2 framework. The left part shows the architecture of learning on image-text pairs, which produces grounded tokens based on the sharing semantics in images and texts. The right part shows the architecture of learning on unpaired images and texts, which produces grounded tokens from image representations or text representations, respectively. As they share the same grounded dictionary, the grounded tokens act as ``anchor points'' to bridge the gap between images and texts.}
	\label{fig:univlp}
\end{figure*}

\section{Related Work}

\paragraph{Vision-Language Pre-training}
Recent years have witnessed rapid progress in vision-and-language pretraining (VLP)~\citep{li2019visualbert, lu2019vilbert, chen2020uniter, li2019unicoder, li2020oscar, yu2020ernie}.
%, which learns general cross-modal representations from massive image-text pairs, and then fine-tunes on various downstream V+L tasks \citep{li2019visualbert, lu2019vilbert, chen2020uniter, li2019unicoder, li2020oscar, yu2020ernie}.
Most existing mainstream VLP models adopt a two-stage training method, which firstly extracts region-based visual features using a pre-trained object detection model, and then combines the derived object-centric region features of images and text embeddings as the input of Transformer \citep{vaswani2017attention} for cross-modal pre-training.
These methods rely heavily on an off-the-shelf object detector like Faster R-CNN \citep{ren2016faster} typically pretrained on the Visual Genome dataset \citep{anderson2018bottom}.
%, which annotates 1,600 object classes and 400 attribute classes \citep{anderson2018bottom}.
%The object detection model is usually trained on specific visual datasets such as the Visual Genome dataset (Krishna et al., 2017), which annotates 1,600 object classes and 400 attribute classes (Anderson et al., 2018).
As the visual representation is not optimized towards a more generic cross-modal understanding and extracting region features with an object detection model is so time-consuming, they face the problems of limited visual expressive power and computation inefficiency, which makes them less scalable.
%There are two mainstream architectures for bridging the cross-modal semantic gap: single-stream architecture and two-stream architecture. The former such as VL-BERT (Suet al., 2019) and UNITER (Chen et al., 2019b) simply concatenates image-region features and text features as input to a single Transformer (Vaswani et al., 2017) network for early fusion in a straightforward manner. In contrast, the latter like LXMERT (Tan and Bansal, 2019) and ERNIE-ViL (Yu et al., 2020) first uses separate Transformer encoders to learn high-level abstraction of image and sentence representation respectively, and then combines the two modalities together with a cross-modal Transformer.
%As these methods rely heavily on expensive region-based visual feature extraction by pre-trained object detector,

Some recent work has also explored VLP without object detection modules \citep{xu2021e2e, kim2021vilt, huang2021seeing, wang2021simvlm}.
They either utilize grid features from pretrained CNNs or patch features following ViT \citep{dosovitskiy2020image}, however they only use limited image-caption pairs for cross-modal pretraining and thus their scalability and performance are limited.
Only a few works have explored utilizing unaligned images and texts for vision-language pre-training, including our previous work UNIMO \citep{li2020unimo} and U-VisualBERT \citep{li2020unsupervised}.
However, they all rely on pre-extraction of region-based visual features or object tags by time-consuming object detection.
How to bridge unpaired visual and textual corpora end-to-end without using object detection remains challenging.

\paragraph{Grounded Learning}
Language grounding is an active field aiming at enriching textual representations with visual information, which has been shown to improve performance on a variety of core NLP tasks \citep{bruni2014multimodal, baroni2016grounding,kiela2017deep}. 
%\citep{fincher2001perceptual, andrews2009integrating}.
%, which is motivated by evidence that human understanding of language is grounded in physical reality and perceptual experience \citep{jones1991object, perfetti1998limits, fincher2001perceptual, andrews2009integrating, riordan2011redundancy}.
%Learning multi-modal representations that ground text-only representations has been shown to improve performance on a variety of core NLP tasks \citep{bruni2014multimodal, baroni2016grounding,kiela2017deep}.
\citet{kiela2017learning} investigate grounded sentence representations by training a sentence encoder to predict the image features of a given caption. %and use the resultant features as sentence representations.
%\citet{bordes2020incorporating} propose to transfer visual information to textual representations by learning an intermediate representation space.
\citet{tan2020vokenization} propose a vokenization method that maps language tokens to their related images.
These works all enrich the language representation with visual information by learning a projection of text representations to corresponding images \citep{chrupala-etal-2015-learning}. 
Recently, \citet{huang2021seeing} propose an end-to-end VLP method that aggregates visual features from a CNN encoder into visual tokens with a visual dictionary. 
\citet{liu2021cross} propose to improve cross-modal retrieval tasks by incorporating a shared discretized embedding space, which is utilized to compute matching scores between different modalities to complement the representations from individual encoders.
These works all rely on image-text pairs to learn cross-modal representations and only focus on joint vision-language tasks.
By contrast, our work for the first time proposes to jointly model both aligned and unaligned images and texts by end-to-end learning a shared grounded semantic space, which can improve modality alignment between both aligned and unaligned images and texts.

\section{Approach}
The overall architecture of our model is shown in Figure \ref{fig:univlp}.
UNIMO-2 is an end-to-end framework, which consists of a trainable Transformer-based visual encoder, a Transformer-based text encoder, a grounded dictionary (GD) embedding module, and a multi-layer Grounded Transformer for modality fusion.
The visual encoder takes an image as input by splitting it into small sizes of patches, and produces the high-level visual representations for all patches, similar to ViT \citep{dosovitskiy2020image}.
The text encoder encodes textual tokens to produce high-level token representations.
Based on the high-level representations of patches and tokens, we design a GD embedding module to group similar vision-language representations into grounded tokens with a shared grounded dictionary.
The Grounded Transformer is further adopted to fuse features from vision and language modalities through interacting with the common grounded tokens.
UNIMO-2 can be end-to-end pre-trained by joint Masked Language Modeling (MLM) on text, Image-Text Matching (ITM) on image-text pairs and Visual Contrastive Learning (VCL) on images.
UNIMO-2 can also be easily adapted to various tasks including visual, textual and cross-modal tasks.

\subsection{End-to-End Grounded Learning}
Human acquire much of their knowledge through grounded learning – visual concepts can be acquired through language, and language acquisition emerges through visual interaction \citep{jones1991object, perfetti1998limits, fincher2001perceptual, andrews2009integrating, riordan2011redundancy}.
Inspired by this type of grounded learning, we propose to learn a sharing semantic space (i.e. grounded space) between images and texts to better align fine-grained visual and textual semantics.
Specifically, based on the high-level visual representations of patches ${V}=\{v_1,\dots,v_M\}$ and textual representations of tokens ${T}=\{t_1,\dots,t_N\}$, we introduce a grounded dictionary to group similar visual and textual representations into the same grounded token.
The grounded features not only help align the visual and textual semantics in aligned image-caption data, but also act as ``anchor points'' to help bridge the unaligned images and texts, as shown in Figure~\ref{fig:univlp}.

\paragraph{Grounded Dictionary Learning}
We define a grounded dictionary (GD) as a matrix $G \in \mathbb{R}^{C \times D}$ which contains $C$ embedding vectors with $D$-dim. The embedding vector for the $j^{th}$ grounded token is denoted as $g_j \in \mathbb{R}^{D}, j \in 1,2,\ldots,C$. 
Vector Quantization (VQ) is widely used to group continuous embeddings into groups of discrete latent variables \citep{oord2017neural, liu2021cross, huang2021seeing}.
For example, each patch or token can be mapped to a grounded token by finding its nearest neighbor in the GD, as in \citet{oord2017neural}.

Most existing VLP methods implicitly assume that there is a one-to-one correspondence hypothesis between the visual and textual modalities of image-text pairs.
However, this hypothesis does not hold in reality as most image-text pairs on the Web are noisy or only have weak correlation.
To tackle this issue, instead of mapping each patch or token representation to a grounded token, we only detect the most significant sharing semantics between image and text.
We propose to find the top-$K$ most significant grounded tokens for both the textual and visual input.
%Instead of searching the mapping index for each patch and token, we select the top-K most significant grounding index for all patches and tokens. 
Specifically, let $x_{ij}$ denotes the similarity between embedding vectors of visual token $v_i$ and grounded token $g_j$, which is computed by:
\begin{equation}
\begin{aligned}
    x_{ij} = \sigma(\eta * v_i^T g_j)
\end{aligned}
\label{eq1}
\end{equation}
\noindent where $\sigma$ denotes the $sigmoid$ function, and $\eta$ denotes a learnable temperature parameter. 
Similarly, $y_{kj}$ denotes the similarity between embedding vectors of textual token $t_k$ and grounded token $g_j$. 

For image-text pairs, the accumulated score of the grounded token $g_j$ is computed as:
\begin{equation}
\begin{aligned}
    s_j = \sum_{i=1}^{M} x_{ij} + \sum_{k=1}^N y_{kj}
\end{aligned}
\label{eq2}
\end{equation}
We obtain the top-$K$ most significant grounded tokens with the largest accumulated scores: $g_1,\ldots,g_K = Top_K \{s_1,\ldots,s_C\}$, where $K$ is a hyper-parameter.
Note that, if we set $K=M+N$, then it is similar that each patch or token is mapped to a grounded token, which will increase the computation cost and introduce noisy information into the grounded learning process.
So, we set $K$ much smaller than $M+N$ to obtain the most significant and sharing grounded tokens, which can help align fine-grained visual and textual representations while eliminating the noisy or unrelated information in image-text pairs.
For unpaired images or text, the accumulated score of each grounded token $g_j$ is $s_j = \sum_{i=1}^{M} x_{ij}$ or $s_j = \sum_{k=1}^N y_{kj}$, and the top-$K$ grounded tokens can be obtained similarly.

The grounded dictionary is randomly initialized, and further updated end-to-end while pre-training. As the $Top_K$ function is non-differentiable, we import a \emph{grounding loss} to help learn the grounded dictionary.
Specifically, we propose a revised form of the Vector Quantisation (VQ) algorithm \citep{oord2017neural}, which uses the $l_2$ error to move the embedding vectors $g_i$ towards the mapped patch or token representations, as shown in the first term of Equation \ref{eq3}.
For simplicity, here we take image input as an example.
Since the volume of the embedding space is dimensionless, it can grow arbitrarily if the embeddings $g_i$ do not train as fast as the visual and textual encoder parameters. To make sure the encoder commits to an embedding and its output does not grow, we add a commitment loss, the second term in Equation \ref{eq3}. 
Thus, the total grounding loss becomes:
\begin{equation}
\begin{aligned}
    \mathcal{L}_{GD} =& \sum_{i=1}^{M} \| sg(v_i) - \sum_{j} \frac{x_{ij}}{\sum_{k} x_{ik}} g_j \|_2^2 \\
    & + \beta \sum_{j=1}^{K} \| sg(g_j) - \sum_{i} \frac{x_{ij}}{s_j} v_i \|_2^2
\end{aligned}
\label{eq3}
\end{equation}
\noindent where $sg(.)$ denotes the stop-gradient operator that is defined as identity at forward computation time and has zero partial derivatives, and $\beta$ denotes a weight parameter.
%Our experiments show that the loss is quite robust to $\beta$, as the results did not vary for values of $\beta$ ranging from $0.1$ to $1.0$. We set $\beta = 0.25$ in all our experiments.
%We found the resulting algorithm to be quite robust to $\beta$, as the results did not vary for values of $\beta$ ranging from $0.1$ to $2.0$. We use $\beta = 0.25$ in all our experiments.

The grounded dictionary faces a cold-start problem for unpaired images and texts. So we apply curriculum learning on different types of corpora. Specifically, we first only train on image-text pairs for 20 epoches to obtain a usable grounded embedding space, then further train on all three types of corpus to help bridge unpaired images and texts. To show what the GD has learned, we have visualized some grounded tokens in Appendix~\ref{sec:grounded_ana}.

\begin{figure}[t!]
	\centering
	\includegraphics[width=2.5in]{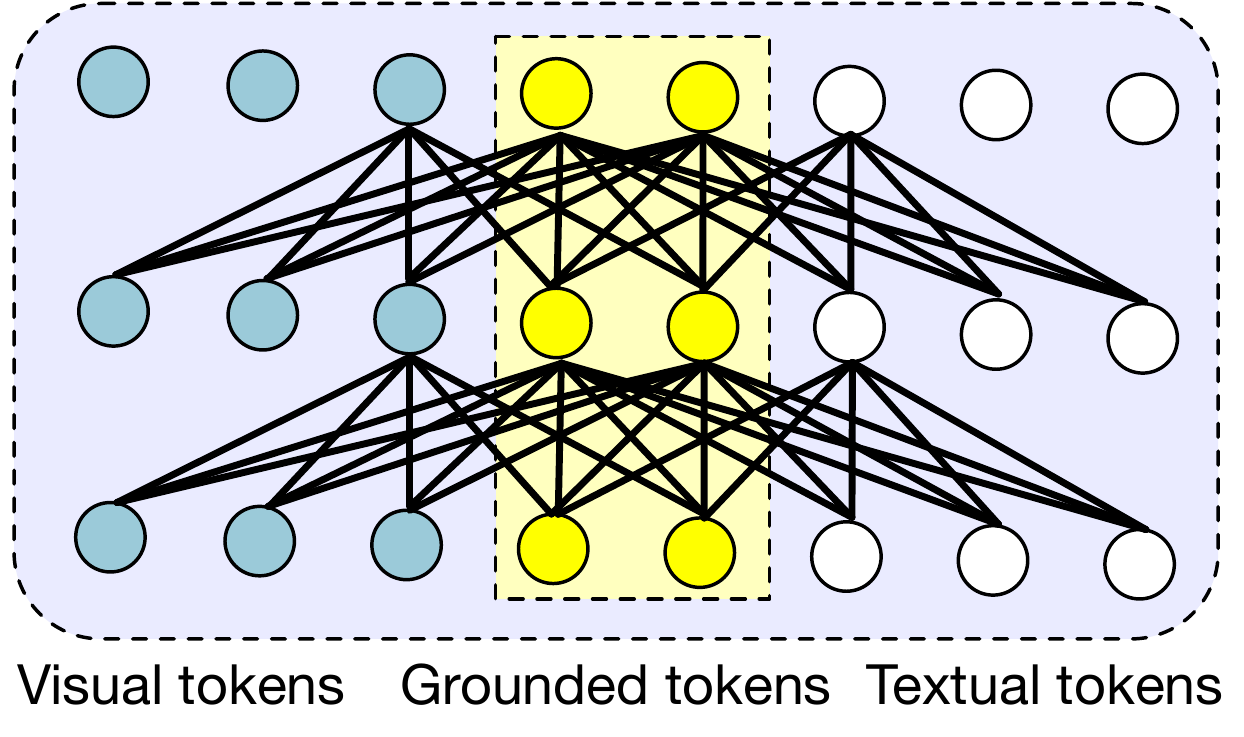}
	\caption{The self-attention architecture of Grounded Transformer. Cross-modal information is exchanged through the grounded tokens.}
	\label{fig:groungedtransformer}
\end{figure}

\paragraph{Grounded Transformer}
After obtaining the grounded tokens, we append them with the visual tokens and textual tokens as input to our Grounded Transformer for cross-modal fusion.
Specifically, we propose to bridge visual and textual representations by grounded tokens.
As shown in Figure \ref{fig:groungedtransformer}, the cross-modal information can only be exchanged by grounded tokens, which also push the grounded tokens to capture the most significant sharing semantics between images and texts.
In this way, our model is more robust on weak correlation image-text pairs by modeling cross-modal interaction through common grounded tokens.
Furthermore, the novel self-attention architecture can improve the computation efficiency compared to the standard pairwise self-attention mechanism.

For unpaired images and texts, the Grounded Transformer also models the fusion of visual tokens or textual tokens with the grounded tokens.
As the grounded dictionary captures common visual and textual semantics, it also helps learn cross-modal representations on unpaired images and texts. 

\subsection{Pre-training On Different Corpus}
Based on the outputs of the Grounded Transformer, we adopt Masked Language Modeling (MLM) and Image-Text Matching (ITM) pre-training tasks on image-text pairs. Furthermore, we also apply MLM on text corpus and Visual Constrastive Learning (VCL) on images.

\paragraph{Masked Language Modeling}
We iteratively sample spans of text until totally 15\% tokens have been selected. We sample the span length from a geometric distribution $l \sim Geo(p)$, where $p$ is set as 0.2, similar to SpanBERT \citep{joshi-etal-2020-spanbert}. All tokens in the selected spans are replaced with either a special [MASK] token, a random token or the original token with probability 80\%, 10\% and 10\%, respectively. 
The goal is to predict these masked tokens based on their surrounding context and all visual features.
%The goal is to predict these masked tokens $t_m$ based on their surrounding context $T_{\setminus m}$ and all visual features $V$, by minimizing the negative log-likelihood:
%\begin{equation}
%\begin{aligned}
%    \mathcal{L}_{MLM} = -\mathbb{E}_{(T,V) \sim D}\ log\ p(t_m | T_{\setminus m}, V)
%\end{aligned}
%\label{eq4}
%\end{equation}
The MLM task is also applied on text-only corpus, which predicts masked tokens only based on the surrounding tokens.

\paragraph{Image-Text Matching}
To enhance the cross-modal matching, we adopt ITM task for pre-training as in previous works \citep{chen2020uniter}. We apply a binary classifier on the concatenated embedding features of the ``[CLS]'' token in text and the ``[CLS]'' token in image by Grounded Transformer to predict whether the input image and text are matched or not. %ITM task is driven by the following loss function:

\paragraph{Visual Contrastive Learning}
UNIMO-2 learns representations on unpaired images by maximizing agreement between differently augmented views of the same image while minimizing similarities between different images via a contrastive loss in the latent space, similar to SimCLR \citep{chen2020simple}.
We apply stochastic data argumentation module that transforms an image randomly resulted in two corelated views as a positive pair, and random images in the same minibatch as negative pairs.
We combine augmentations of random cropping, random rotating and random color distortion followed by resizing back to the original size.
%The contrastive loss function for a positive pair of $v_i$ and $v_j$ is denfined as:
%\begin{equation}
%\begin{aligned}
%    \mathcal{L}_{VCL} = -log \frac{exp(sim(v_i, v_j)/\tau)}{\sum_k \mathbb{I}_{[k \ne i]} exp(sim(v_i, v_k)/ \tau) }
%\end{aligned}
%\label{eq7}
%\end{equation}
%\noindent where $\mathbb{I} \in {0,1}$ is an indicator function evaluating to 1 if $k \ne i$ and $\tau$ denotes a learnable temperature parameter.

%The visual encoder, text encoder, grounded embedding component and the Grounded Transformer are end-to-end jointly trained on all three types of corpus.
%The full pretraining objective of UniVLP is:
%\begin{equation}
%\begin{aligned}
%    \mathcal{L} = \mathcal{L}_{GD} + \mathcal{L}_{MLM} + \mathcal{L}_{ITM} + \mathcal{L}_{ITCL} + \mathcal{L}_{VCL}
%\end{aligned}
%\label{eq7}
%\end{equation}

\subsection{Transferring To Different Tasks}
Our model can be effectively finetuned on different types of tasks, including cross-modal tasks, visual tasks and textual tasks.
For cross-modal tasks, the model architecture is the same as the pre-training architecture on image-text pairs, as shown in the left part of Figure \ref{fig:univlp}.
Grounded tokens are produced based on both the visual and textual representations to facilitate cross-modal understanding and generation.
%Experiments show that UniVLP has very good zero-shot performance on cross-modal image-text/text-image retrieval.
For visual tasks, the model architecture is the same as the pre-training architecture on images, as shown in the middle part of Figure \ref{fig:univlp}.
Grounded tokens are obtained based on the visual representations from the Visual Transformer.
As the grounded tokens contain sharing semantics between images and texts, UNIMO-2 can learn language-grounded image representations for visual tasks.
Similarly, for textual tasks the model architecture is the same as the pre-training architecture on text, as shown in the right part of Figure \ref{fig:univlp}.
Grounded tokens are obtained based on the textual representations from the Text Transformer.
Also, the sharing grounded space helps learn grounded text representations to facilitate textual tasks.

\section{Experimental Settings}
%In this section, we introduce our pre-training and finetuning experimental settings.

\paragraph{Pretraining Dataset}
Our pre-training datasets consist of three types: text corpus, image corpus and image-text pairs. 
The text corpus includes two large-scale corpora: BookWiki and OpenWebText, which are part of the training dataset of RoBERTa \citep{liu2019roberta}. 
%BookWiki is composed of English Wikipedia and BookCorpus (Zhu et al., 2015), and OpenWebText is an open recreation of the WebText corpora. 
The image corpus are images without textual descriptions, including a subset of OpenImages \citep{krasin2017openimages} and ImageNet-21k \citep{deng2009imagenet}. Each image in these datasets contains a textual label. 
The image-text pairs are composed of four existing multi-modal datasets: COCO \citep{lin2014microsoft}, Visual Genome (VG) \citep{krishna2017visual}, Conceptual Captions (CC) \citep{sharma-etal-2018-conceptual} and SBU Captions \citep{ordonez2011im2text}, which have also been widely used in previous VLP models.
The detail statistics are shown in the appendix.
We also transform the label of each image to a sentence by prompts (e.g. ``a photo of [label]'') to create pseudo image-text pairs from the OpenImages and ImageNet-21k datasets for pretraining.
%The number of image-text pairs is 9.5M.
%To show that our method is scalable with noisy web data or weak correlation data, we also include the much noisier Conceptual 12M (cc12m) dataset \citep{changpinyo2021conceptual}, increasing the total number of pairs to 17M (part of cc12m pairs are lost).

\begin{table*}[ht!]
\centering
\small
\begin{tabular}{l c c c c c c c}
\hline \hline
\multirow{2}{*}{Model} & ZS-IR & ZS-TR & IR & TR & SNLI-VE & VQA & Caption \\
& R@1/R@5 & R@1/R@5  & R@1/R@5  & R@1/R@5 & Val / Test & test-dev / std  & B@4 / C \\
\hline
\multicolumn{8}{l}{\emph{Region-based Models Pretrained on Image-Text Pairs of CC, SBU, COCO and VG.}} \\
%VLP-Base & - & - & - & - & - & 70.5/70.7 & 36.5/116.9 \\
ViLBERT & 31.86/61.12 & - & 58.20/84.90 & - & - & 70.55/70.92 & - \\
UNITER-Base & 66.16/88.40 & 80.70/95.70 & 72.52/92.36 & 85.90/97.10 & 78.59/78.28 & 72.70/72.91 & - \\
Villa-Base & - & - & 74.74/92.86 & 86.60/97.90 & 79.47/79.03 & 73.59/73.67 & -\\
Oscar-Base & - & - & - & - & - & 73.16/73.44 & 36.5/123.7 \\
UNIMO-Base & 62.44/86.16 & 77.40/95.10 & 74.66/93.40 & 89.70/98.40 & 80.00/79.10 & 73.79/74.02 & 38.8/124.4 \\
UNITER-Large & 68.74/89.20 & 83.60/95.70 & 75.56/94.08 & 87.30/98.00 & 79.39/79.38 & 73.82/74.02 &  -  \\
Villa-Large & - & - & 76.26/94.24 & 87.90/97.50 & 80.18/80.02 & 74.69/74.87 & - \\
Oscar-Large & - & - & - & - & - & 73.61/73.82 & 37.4/127.8 \\
%ERNIE-ViL-large & - & - & 76.70 & 88.10 & - & 74.75 / 74.93 & - \\
%ALBEF & 82.8 & 94.1 & 85.6 & 95.9 & 80.80 / 80.91 & 75.84 / 76.04 & - \\
UNIMO-Large & 72.14/91.14 & 85.80/96.80 & 78.04/94.24 & 89.40/98.90 & 81.11/80.63 & 75.06/75.27 & 39.6/127.7 \\
\hline
\multicolumn{8}{l}{\emph{End-to-End Models Pretrained on Image-Text Pairs of CC, SBU, COCO and VG. $\dagger$ denotes 400 Million pairs.}} \\
%Ernie-ViL-base & - & - & 74.44 & 86.70 & - & 72.62 / 72.85 &  - \\
ViLT & 51.3/79.9 & 69.7/91.0 & 62.2/87.6 & 83.7/97.2 & - & 70.94/- & - \\
E2E-VLP & - & - & 73.58/92.42 & 86.24/97.50 & - & 73.25/73.67 & 36.2/117.3 \\
SOHO & - & - & 72.5/92.7 & 86.5/98.1 & \textbf{85.00}/\textbf{84.95} & 73.25/73.47 & - \\
CLIP$\dagger$ & 68.7/90.6 & 88.0/\textbf{98.7} & - & - & - & - & - \\
\rowcolor[gray]{.9}
\textbf{Our Baseline} & 65.11/87.44 & 78.80/94.38 & 78.52/94.02 & 91.62/98.72 & 80.37/80.43 & 75.69/75.87 & 38.5/128.4 \\
\rowcolor[gray]{.9}
\textbf{UNIMO-2} & \textbf{72.70}/\textbf{91.18} & \textbf{88.46}/96.84 & \textbf{80.14}/\textbf{95.58} & \textbf{92.01}/\textbf{99.31} & 81.97/81.48 & \textbf{76.31}/\textbf{76.42} & \textbf{39.7}/\textbf{131.2} \\
\hline
%\rowcolor[gray]{.9}
%\textbf{UniVLP-large} &  &  &  &  & & & \\
\hline
 \hline
\end{tabular}
\caption{\label{cross-modal}
Evaluation results on cross-modal tasks. ZS denotes zero-shot performance. IR and TR represents image-retrieval and text-retrieval, respectively. B@4 and C denotes metrics of BLUE4 and CIDEr, respectively. ``Our Baseline'' is similar to UNIMO-2, except that the grounded embedding module in UNIMO-2 is removed. It is trained on the same corpus and experimental settings with UNIMO-2.}
\end{table*}

\begin{table}[t!]
\centering
\small
\begin{tabular}{l c c}
\hline \hline
\multirow{2}{*}{Model} & \multicolumn{2}{c}{Acc@1}  \\
 & Zero-Shot & Finetuned \\
\hline
SimCLRv2 \citep{chen2020big} & - & 80.5 \\
%ViT-B/16(ImageNet-21k) & - & \textbf{84.0} \\
CLIP-ViT(B/16) & \textbf{68.6} & 80.2 \\
%\hline
\rowcolor[gray]{.9}
\textbf{Our Baseline} & 58.2 & 80.7  \\
\rowcolor[gray]{.9}
\textbf{UNIMO-2} & 66.3 & \textbf{80.8} \\
%\hline
%ViT-L/16(ImageNet-21k) & - & 79.5 \\
%CLIP-ViT(L/14-336px) & 76.2 & 85.4 \\
%\rowcolor[gray]{.9}
%\textbf{UniVLP-large} &  & \\
\hline
\hline
\end{tabular}
\caption{\label{visual}
Evaluation results on visual tasks, compared to state-of-the-art representation learning methods. We report both the zero-shot and finetuned top-1 accuracy on ImageNet-1k. The finetuned result of CLIP-ViT is linear probe performance.}
\end{table}

\paragraph{Implementation Detail}
%We evaluate UniVLP on two model sizes: UniVLP-large and UniVLP-Base.
%UniVLP-large consists of 24 layers of Visual Transformer, 12 layers of Text Transformer and 12 layers of Grounded Transformer.
%The Visual Transformer is initialized by ViT-L/16.
%The Text Transformer and Grounded Transformer are initialized by the first 12 layers and last 12 layers of RoBERT-Large, respectively. 
%Similarly, UniVLP-base are initialized by Vit-B/16 and RoBERTa-Base.
UNIMO-2 consists of 12 layers of Visual Transformer, 12 layers of Text Transformer, and 12 layers of Grounded Transformer.
The Visual Transformer is initialized by ViT-B/16.
The Text Transformer and Grounded Transformer are both initialized by RoBERTa-Base. 
The maximum sequence length of text tokens are set as 512.
An Adam optimizer with initial learning rate 5e-5 and a learning rate linear decay schedule is utilized.

For the visual encoder, 
%we utilize the $224 \times 224$ resolution with a fixed patch size of $16 \times 16$ during pre-training. During fine-tuning, we increase the image resolution to $384 \times 384$ and interpolate the positional encoding of image patches following \citep{dosovitskiy2020image}.
our model receives the raw image $\mathbf{x} \in \mathbb{R}^{H \times W \times C}$ and maps it into flattened $1D$ sequence of patches $\mathbf{x}_p \in \mathbb{R}^{\frac{HW}{P^2} \times D}$ as input for the transformer, where $D$ is the fixed hidden size of the transformer layers and $P$ is the patch size. During pretraining, we utilize the $224 \times 224$ resolution with a fixed patch size of $16 \times 16$, resulting in a patch sequence of length $14 \times 14$ as visual tokens. During fine-tuning, we increase the image resolution to $384 \times 384$ and interpolate the positional encoding of image patches following \citep{dosovitskiy2020image}.
For the grounded embedding module, the grounded dictionary size $C$ is set as 2048, and the number of grounded tokens $K$ during pre-training and finetuning are both set as 100 that is much smaller than the max number of patches and tokens for pre-training (i.e. 709) and finetuning (i.e. 1089). We set $\beta = 0.25$ in all our experiments and the results did not vary obviously for values ranging from $0.1$ to $1.0$. We have compared different grounding settings in detail in Appendix~\ref{sec:grounded_ana}.

\paragraph{Finetuning Tasks}
To show the scalability of our model, we fine-tune it on three types of downstream tasks: (1) joint vision-language cross-modal tasks, (2) visual tasks, and (3) textual tasks. The cross-modal tasks include: visual question answering (VQA) on the VQA v2.0 dataset \citep{goyal2017making}, image caption on the Microsoft COCO Captions dataset \citep{chen2015microsoft}, visual entailment on the SNLI-VE dataset \citep{xie2019visual} and image-text retrieval on Flickr30k datasets \citep{young2014image}. The visual tasks include image classification on the ImageNet-1k dataset \citep{krizhevsky2012imagenet}. The textual tasks include sentiment classification on the SST-2 dataset \citep{socher-etal-2013-recursive}, natural language inference on the MNLI dataset \citep{williams2017broad}, linguistic acceptability analysis on the CoLA dataset \citep{warstadt2019cola} and semantic similarity analysis on the STS-B dataset \citep{cer-etal-2017-semeval}.  The detail statistics of the datasets and hyper-parameter settings for the above tasks are described in Appendix~\ref{sec:detail_settings}.

\section{Results and Analysis}
We compare UNIMO-2 to a variety of state-of-the-art models on cross-modal, visual and textual tasks.% to show its adaptability and scalability. % to different tasks. 
%We further make several ablation studies to validate effectiveness of our grounded learning method in jointly aligning both paired and unpaired images and text. We will also do extensive parameter analysis for grounded learning to show some properties of the grounded space.

\subsection{Cross-Modal Tasks}
The evaluation results on the joint vision-language cross-modal tasks are shown in Table \ref{cross-modal}.
%ViLBERT \citep{lu2019vilbert}, VLP \citep{zhou2020unified},
We compare with most of the existed VLP models, including regional feature-based models ViLBERT~\citep{lu2019vilbert}, UNITER~\citep{chen2020uniter}, Oscar~\citep{li2020oscar}, Villa~\citep{gan2020large} and UNIMO~\citep{li2020unimo}, and end-to-end models ViLT~\citep{kim2021vilt}, E2E-VLP~\citep{xu2021e2e}, SOHO~\citep{huang2021seeing} and CLIP~\citep{radford2021learning}.% and ALBEF \citep{li2021align}. 
The results show that UNIMO-2 achieves the best results against most benchmarks, outperforming both the base and large sizes of other VLP models.
Particularly, UNIMO-2 achieves very good performance on the task of zero-shot image/text retrieval, even outperforming CLIP \citep{radford2021learning} that pre-trained on an order of magnitude larger corpus. The results demonstrate that UNIMO-2 can obtain better cross-modal representations based on joint end-to-end grounded learning on different types of corpus.

%UniVLP achieves better performance on both the cross-modal understanding and generation tasks, while most previous methods usually focus on either the understanding (i.e. Image/Text Retrieval, SNLI-VE and VQA) or generation tasks (i.e. CoCo Caption).
Furthermore, the performance of ``Our Baseline'' that just removes the grounded embedding module in UNIMO-2 drop obviously on all tasks, which demonstrates the effectiveness of our grounded learning method for cross-modal alignment.
Especially, on the zero-shot image retrieval and text retrieval tasks, UNIMO-2 obtains 7.59 R@1 and 9.66 R@1 absolute gains compared to ``Our Baseline''. 
The results demonstrate that our grounded learning method can help align the visual and textual semantic space on different types of corpora to obtain more effective cross-modal representations.
%On the image caption task, UNIMO outperforms the best performing model Oscar by more than 2 BLUE4 score.
%On the image caption task, UNIMO outperforms the best performing model Oscar by 2.1 BLUE4 score.
%On the visual entailment, UNIMO also outperforms all previous models and achieves the best performance.
%The above results also demonstrate the effectiveness of our unified end-to-end VLP architecture that jointly learns on both aligned and unaligned images and texts for cross-modal learning.

\begin{table}[t!]
 \centering
 \setlength{\tabcolsep}{2.7pt}
 \small
 \begin{tabular}{l c c c c c c c c}
  \hline \hline
  \multirow{2}{*}{Model} & SST-2 & MNLI & CoLA & STS-B\\
  & Acc & Acc-(m/mm)& Mat & Per \\
  \hline
  BERT & 92.7 & 84.4 / - & - & -  \\
  RoBERTa & 94.8 & - & 63.6 & - \\
  UniLM & 94.5 & 87.0/85.9 &  61.1 & 87.7 \\
  UNITER & 89.7 & 80.8/- & 37.4 & - \\
  VilBERT & 90.4 & 79.9/- & 36.1 & - \\
  UNIMO & \textbf{95.1} & 86.8/86.7 & \textbf{65.4} & 91.0 \\
  \rowcolor[gray]{.9}
  \textbf{Our Baseline} & 94.1 & 87.1/86.9 & 60.6 & 91.0 \\
  \rowcolor[gray]{.9}
  \textbf{UNIMO-2} & 94.7 & \textbf{87.5}/\textbf{87.5} & 62.1 & \textbf{91.2} \\
  %\hline
  %BERT-large & 93.2 & 86.6/- & 60.6 & 90.0 \\
  %RoBERTa-large & 96.4 & \textbf{90.2}/\textbf{90.2} & 68.0 & 92.4 \\
  %XLNet-large & 95.6 & 89.8/- & 63.6 & 91.8 \\
  %%UniLM-large & 94.5 & 87.0/85.9 &  61.1 & 87.7 \\
  %UNIMO-large & 96.8 & 89.8/89.5 & 68.5 & 92.6 \\
  %\rowcolor[gray]{.9}
  %\textbf{UniVLP-large} & \textbf{} & & \textbf{} & \textbf{} \\
  \hline
  \hline
 \end{tabular}
 \caption{\label{textual}
 %Evaluation results on the single-modal downstream tasks.
 Evaluation results on textual tasks. Mat and Per denote Matthews correlation coefficient and Pearson correlation coefficient, respectively. All the results are evaluated on the dev set.}
\end{table}

\begin{table*}[t!]
\centering
\small
\begin{tabular}{c l c c c c c c c}
\hline \hline
& \multirow{2}{*}{Model} & ZS-IR & ZS-TR & IR & TR & COCO Caption & ZS-ImageNet & MNLI \\
& & R@1 & R@1 & R@1 & R@1 & B@4 / C & Acc@1 & m/mm \\
\hline
\rowcolor[gray]{.9}
 & \textbf{UNIMO-2} & \textbf{72.70} & \textbf{88.46} & \textbf{80.14} & \textbf{92.01} & \textbf{39.7} / \textbf{131.2} & 66.3 & \textbf{87.5}/\textbf{87.5}\\
%\hline
\multirow{4}{*}{GD} & w/o GD (P) & 65.11 & 78.80 & 78.52 & 91.62 & 38.5 / 128.4 & 58.2 & 87.1/86.9\\
& w/o GD (I) & 40.22 & 31.76 & 74.08 & 88.26 & 39.0 / 127.4 & 21.3 & 87.5/87.3 \\
& w/o G.T. & 70.10 & 85.01 & 78.84 & 91.12 & 39.6 / 130.1 & \textbf{66.4} & 87.1/86.8\\
& 1-to-1 Map & 66.06 & 80.97 & 77.61 & 90.43 & 38.7 / 127.4 & 66.3 & 87.0/86.9 \\
\hline
\multirow{3}{*}{Corpus} & w/o Text & 70.00 & 85.50 & 78.90 & 90.24 & 39.0 / 128.7 & 65.0 & 84.9/85.0\\
& w/o Images & 69.17 & 84.81 & 77.65 & 90.34 & 39.4 / 129.5 & 42.2 & 87.1/87.0\\
& w/o Both & 70.06 & 84.12 & 78.17 & 91.32 & 39.3 / 129.3 & 43.0 & 85.9/85.7\\
\hline
\hline
\end{tabular}
\caption{\label{ablation}
Ablation study on the effectiveness of our unified end-to-end grounded learning architecture. }
\end{table*}

\subsection{Visual Tasks}
UNIMO-2 can also be effectively adapted to visual tasks such as image classification.
As UNIMO-2 learns effective cross-modal representations, it can classify images without finetuning.
Specifically, the target labels of images can be transformed into pseudo image descriptions, such as ``a photo of [label]''. 
Then the zero-shot image-to-text retrieval method can be used to obtain the label for each image, similar to CLIP \citep{radford2021learning}.
Both the zero-shot and finetuned performance is compared to several state-of-the-art representation learning methods.
The results in Table \ref{visual} show that UNIMO-2 can achieve comparable performance with CLIP that pretrained on billions of image-text pairs, on both the zero-shot and supervised settings.
Moreover, UNIMO-2 obviously outperforms ``Our Baseline'' on the zero-shot setting, achieving 8.1 Acc@1 absolute gains.
The results demonstrate that UNIMO-2 also learns generalized visual representations through unified-modal learning on different types of corpora.

\subsection{Textual Tasks}

To show the effectiveness of UNIMO-2 on textual tasks, we further compare with both VLP models including UNITER, VilBERT and UNIMO, and pre-trained language models including BERT~\citep{devlin-etal-2019-bert}, RoBERTa~\citep{liu2019roberta} and UniLM~\citep{dong2019unified}. 
The comparison results in Table \ref{textual} demonstrate that UNIMO-2 achieves much better performance than existing VLP models including UNITER and VilBERT, and achieves comparable performance than existed PLMs such as RoBERTa.
UNIMO-2 also outperforms ``Our Baseline'' on all textual tasks.

The above results demonstrate the adaptability and scalability of our unified end-to-end VLP architecture for joint learning on both aligned and unaligned images and texts. In all, UNIMO-2 not only achieves excellent performance on cross-modal tasks, but also performs very well on visual and textual tasks, which validates the superiority of our unified-modal learning architecture.

\subsection{Analysis}
\label{sec:analysis}

\paragraph{Effectiveness of Grounded Learning}
We further validate the effectiveness of our grounded learning component by ablation study.
``w/o GD (P)'' denotes removing the grounded learning component during both pre-training and inference in order to validate its effectiveness for unified learning on different types of corpus.
``w/o GD (I)'' denotes keeping the grounded learning component during pre-training, but removing it during inference, in order to validate the effectiveness of the grounded representations to downstream tasks.
``1-to-1 Map'' denotes mapping each patch or token to a grounded token by finding its nearest neighbor in the grounded dictionary, similar to the vector quantization method in \citep{oord2017neural}.
We compare their performance on three types of tasks, as shown in the top part of Table~\ref{ablation}.
The results demonstrate that our grounded learning (GD) method is essential to the end-to-end joint learning from different types of corpus, which can help bridge unaligned images and texts and improve vision-language semantic alignment. 
The learned grounded representations is also critical to both the cross-modal and single-modal downstream tasks.
We further validate the effectiveness of our Grounded Transformer by replacing it with a traditional Transformer, denoted as ``w/o G.T.''.
The results show that the performance of cross-modal tasks drop obviously compared to UNIMO-2, which demonstrate the effectiveness of our Grounded Transformer architecture. %to grounded learning.

\paragraph{Effectiveness of Unaligned Images and Texts}
To further validate the effectiveness of unaligned images and texts to cross-modal learning, we compare the performance of UNIMO-2 on different pre-training datasets.
Specifically, we compare the performance of UNIMO-2 by either removing the text cropus (i.e. ``w/o Text''), the image corpus (i.e. ``w/o Images'') or removing them both (i.e. ``w/o Both'').
The comparison results are shown in the bottom part of Table~\ref{ablation}, which show that either removing text corpus or image corpus will consistently reduce the performance of all three types of tasks, including cross-modal, visual and textual tasks.
It is worth noting that the performance of the image/text retrieval tasks drop obviously when either removing the text-only cropus or image-only corpus, which demonstrate that unaligned corpus is also useful to cross-modal tasks. 
UNIMO-2 can effectively leverage unaligned images and texts to improve cross-modal learning.
%It is worth noting that the performance of the visual task drop obviously when removing the text cropus, and the performance of the textual task also drop when removing the image corpus.
%The results reveal that with our unified learning architecture, the text corpus can enhance the visual task by making the model learn more generalized visual representations, and vice versa.

%\paragraph{Parameter Analysis for Grounded Learning}
%\section{Grounded Representation Analysis}

\section{Conclusion}
In this work, we propose UNIMO-2, an end-to-end unified-modal pre-training framework that can learn from both aligned and unaligned image and text corpora.
Our proposed grounded learning method can help bridge unpaired images and texts and align the textual and visual semantic spaces more effectively.
Benefiting from effectively utilizing different types of corpora, UNIMO-2 has better scalability for different types of tasks. %, including cross-modal, visual and textual tasks.
Experiments show that UNIMO-2 greatly improves the performance of various cross-modal tasks and also achieves very impressive performance on visual and textual tasks.
The results also show that it is promising to further uniformly improve the performance of cross-modal, visual and textual tasks by utilizing larger scales of unpaired images and texts.

\section*{Acknowledgments}
This work was supported in part by the National Key R\&D Program of China under Grant 2020YFB1406701. Xinyan Xiao is the corresponding author.

\bibliography{acl2022}
\bibliographystyle{acl_natbib}

\appendix
\newpage

\begin{table*}[!htbp]
\centering
\small
\begin{tabular}{l c c c c}
\hline \hline
\multicolumn{2}{c}{Model} & ZeroShot-IR & ZeroShot-TR & COCO Caption  \\
& & R@1 / R@5 / R@10 & R@1 / R@5 / R@10 & B@4 / M / C / S \\
\hline
\multirow{4}{*}{GD Size $C$} & 1024 & 58.52 / 82.19 / 88.92 & 71.10 / 90.14 / 95.17 & 37.58 / 29.18 / 123.53 / 22.23 \\
&2048 & 60.32 / 84.02 / 89.72 & 75.84 / 91.91 / 95.56 & 37.62 / 29.12 / 123.38 / 22.16 \\
&4096 & \textbf{64.10} / \textbf{86.41} / \textbf{91.79} & \textbf{77.91} / \textbf{94.38} / \textbf{96.75} & \textbf{38.07} / \textbf{29.20} / 124.18 / 22.20 \\
&8192 & 61.20 / 85.29 / 90.73 & 75.84 / 92.50 / 96.15 & 37.86 / 29.03 / \textbf{124.23} / \textbf{22.33} \\
\hline
\multirow{5}{*}{Top-$K$} & 10 & 57.79 / 82.66 / 89.47 & 69.92 / 91.42 / 95.56 & 37.36 / 28.92 / 122.93 / 22.15 \\
&20 & 61.46 / 85.07 / 90.75 & 74.46 / 93.10 / \textbf{97.34} & 37.90 / 28.81 / 123.68 / 22.03 \\
&50 & \textbf{63.49} / \textbf{86.13} / \textbf{91.54} & \textbf{77.32} / \textbf{93.10} / 96.65 & \textbf{38.38} / \textbf{29.17} / \textbf{125.31} / \textbf{22.39} \\
&100 & 60.32 / 84.02 / 89.72 & 75.84 / 91.91 / 95.56 & 37.62 / 29.12 / 123.38 / 22.16 \\
\hline
& 1-to-1 Map & 56.51 / 81.54 / 88.19 & 71.99 / 90.43 / 94.58 & 35.62 / 27.97 / 117.92 / 21.38 \\
\hline
\hline
\end{tabular}
\caption{\label{parameter}
Parameter analysis for grounded learning. The top part validates the influence of GD size $C$, and the middle part compares the performance of different number of grounded tokens $K$ used during learning. The bottom part shows the effectiveness of our grounded learning method compared with the existing VQ method.}
\end{table*}

\begin{figure*}[!htbp]
	\centering
	\includegraphics[width=5.8 in]{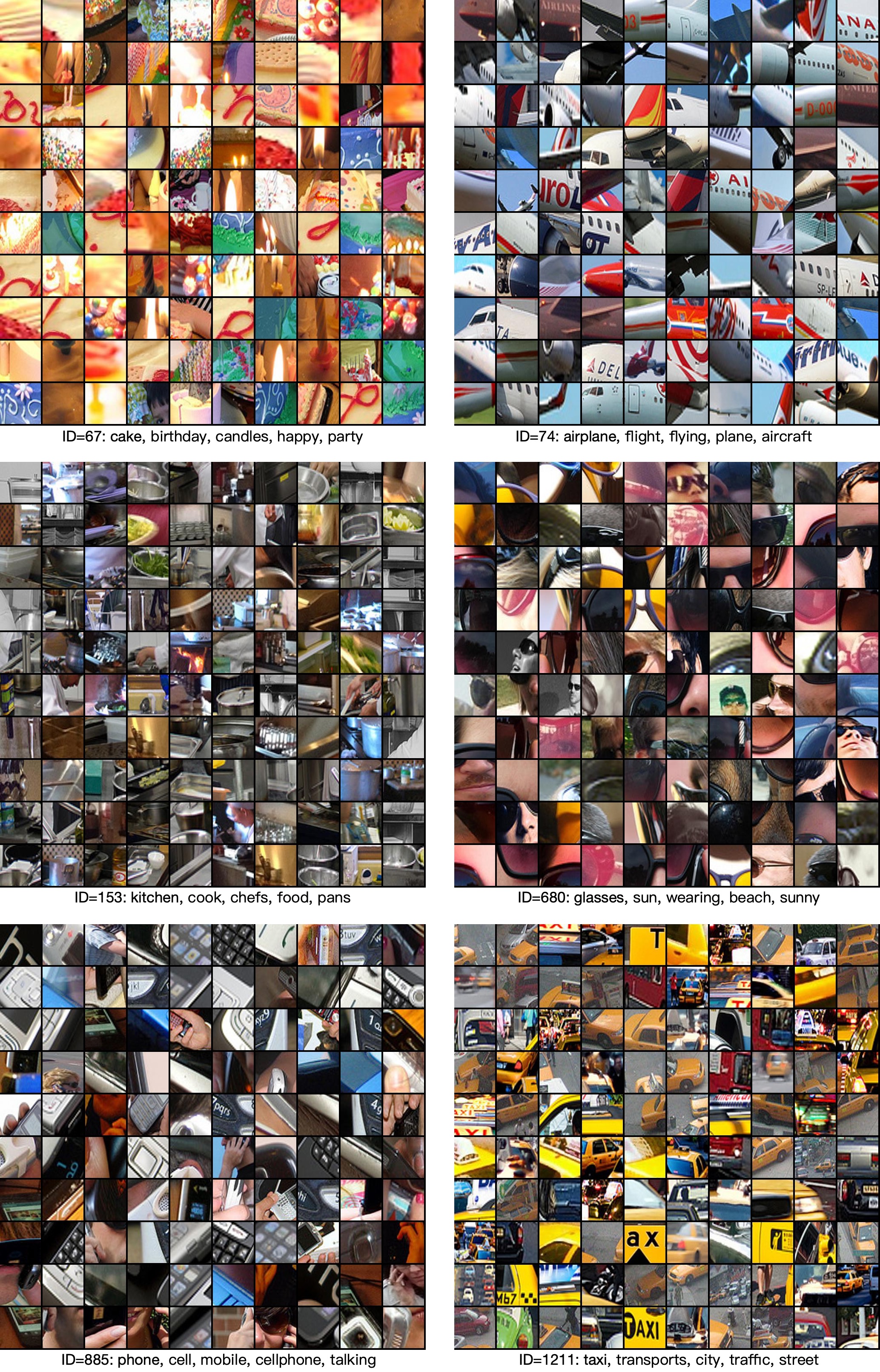}
	\caption{Visualization of the grounded dictionary learned by UNIMO-2, which groups consistent semantics of image patches and textual tokens. Each grounded token reflects an abstraction of vision-language grounded semantics.}
	\label{fig:cases}
\end{figure*}

\section{Grounded Learning Analysis}
\label{sec:grounded_ana}

\paragraph{Visualization of Grounded Dictionary}
To show the semantics of the grounded dictionary learned by UNIMO-2, we visualize the image patches and textual tokens that are grouped in each grounded token. We map each patch or token into a grounded token with which has the largest similarity between their representations by Equation~\ref{eq1}. For each grounded token, the patches and tokens that have the largest similarity scores are selected and visualized. Several examples are shown in Figure~\ref{fig:cases}, which demonstrate that each grounded token captures meaningful and consistent vision-language grounded semantics.

\paragraph{Parameter Analysis}
In all our experiments, we utilize the default grounding settings that the grounded dictionary (GD) size $C$ is set as 2048 and the number of grounded tokens $K$ is set as 100.
We further compare different grounding settings to explore the properties of the grounded semantic space for cross-modal learning.
Specifically, we validate the performance of grounded learning with different grounded dictionary (GD) size $C$ from $\{1024, 2048, 4096, 8192\}$ and different number of grounded tokens $K$ from $\{10, 20, 50, 100\}$.
When comparing different GD size $C$, we set $K$ as 100.
We also keep $C=2048$ when comparing different settings of $K$.
Furthermore, we also compare our method with the simplest Vector Quantization (VQ) method that maps each visual or textual token to a grounded token by finding its nearest neighbor in the grounded dictionary, namely ``1-to-1 map''.
The number of grounded tokens for ``1-to-1 map'' is depended on the total number of image patches and textual tokens, which is 709 (i.e. 197 + 512) during pre-training and 1089 (i.e. 577 + 512) during finetuning.

\begin{table*}[t!]
\centering
\small
\begin{tabular}{l|c|c|c|c|c|c|c|c}
\hline
Type& \multicolumn{4}{c|}{Image-Text Pairs} & \multicolumn{2}{c|}{Unaligned Images} & \multicolumn{2}{c}{Unaligned Text} \\
\cline{1-9}
Dataset & COCO & VG & CC & SBU & ImageNet21K & Open Images & BookWiki & OpenWebText \\
\hline
\#Images & 113K & 108K & 3.01M & 867K & 14M & 1.7M &  &  \\
\#Texts & 567K & 5.41M & 3.01M & 867K & & & 16G & 38G \\
\hline
\end{tabular}
\caption{\label{pretrain-dataset}
Statistics of the aligned image-text pairs, and unaligned images and texts for pre-training.}
\end{table*}

\begin{table*}[!htbp]
\centering
\small
\begin{tabular}{l|l|l|l|l|l}
\hline
\multirow{3}{*}{Task} & \multirow{3}{*}{Image Src.} & \multicolumn{4}{c}{\#Images (\#Text)} \\
\cline{3-6}
&& \multirow{2}{*}{Train} & \multirow{2}{*}{Val} & \multicolumn{2}{c}{Test} \\
\cline{5-6}
&&&& test-std & test-dev \\
\hline
VQA & COCO & 83K(444K) & 41K(214K) & 81K(107K) & 81K(448K) \\
\hline
Image Caption & COCO & 113.2K & 5K & 5K & - \\
\hline
Visual Entailment & Flickr30K & 529.5K & 17.9K & 17.9K & - \\
\hline
Image-Text Retrieval & Flickr30K & 29K(145K) & 1K(5K) & 1K(5K) & - \\
\hline
\end{tabular}
\caption{\label{finetune-dataset}
Statistics of the datasets for the cross-modal downstream tasks.}
\end{table*}

\begin{table}
\centering
\small
\begin{tabular}{l|c|c}
\hline
Hyper-params & Textual Tasks & Visual Tasks \\
\hline
Learning Rate & \{1e-5, 2e-5, 3e-5\} & \{1e-4, 3e-4, 5e-4\} \\
Batch Size & \{16, 32\} & 512 \\
Epochs & 10 & 10 \\
Warmup Raito & 0.06 & 0.06 \\
Weight Decay & 0.01 & 0.01  \\
\hline
\end{tabular}
\caption{\label{gen-params}
Hyper-parameters for fine-tuning on visual and textual tasks.}
\end{table}

\begin{table*}
\centering
\small
\begin{tabular}{l|l|l|l|l}
\hline
Hyper-parameters & Image-Text Retrieval & SNLI-VE & VQA & COCO Caption \\
\hline
\hline
Batch Size & 32 & 64 & 256 & 32 \\
\hline
Epoch & 40 & 10 & 12 & 10 \\
\hline
\multirow{3}{*}{Learning Rate} & 5e-6 for epoch=[0,24] & \multirow{3}{*}{1e-5} & 4e-5 for epoch=[0,5] & \multirow{3}{*}{1e-5} \\
& 5e-7 for epoch=[24,32] & & 4e-6 for epoch=[6,8] & \\
& 5e-8 for epoch=[32,40] & & 4e-7 for epoch=[9,12] & \\
\hline
Warmup Ratio & - & 0.06 & - & 0.06 \\
\hline
Weight Decay & 0.01 & 0.0 & 0.0 & 0.01 \\
\hline
\end{tabular}
\caption{\label{multimodal-params}
Hyper-parameters for fine-tuning on cross-modal tasks .}
\end{table*}

For time efficiency, we only pre-train UNIMO-2 on the corpus of image-text pairs for 10 epoches under the above settings, and then compare their performance on two representative cross-modal tasks, including zero-shot image/text retrieval and image caption, to validate their effectiveness on cross-modal alignment.
The comparison results are shown in Table~\ref{parameter}, which demonstrate that our grounded learning method achieves better performance on the two representative cross-modal tasks when the GD size $C$ is set as 4096 or the number of grounded tokens $K$ is set as 50.
Too large $C$ will increase the difficulty of learning while too small $C$ may restrict the volume of grounded semantic space.
Similarly, too small $K$ will lose sharing semantics between images and texts while too large $K$ will introduce noisy information.
Although different settings have different behavior, the performance of our grounded learning method is relatively stable.
In particular, the ``1-to-1 map'' method achieves much worse results than our grounded learning method under different settings, which validates the effectiveness of our grounded learning method on cross-modal alignment.
Furthermore, our grounded learning method is much more efficient in computation than ``1-to-1 map'' as the number of grounded tokens is much smaller, which largely reduce the sequence length during cross-modal fusion.

\begin{table*}
\centering
\small
\begin{tabular}{l c c c c}
\hline \hline
\multirow{3}{*}{Model} & ZeroShot-IR & ZeroShot-TR & Finetuned-IR & Finetuned-TR  \\
%\cline{2-6}
& R@1 / R@5 / R@10 & R@1 / R@5 / R@10 & R@1 / R@5 / R@10  & R@1 / R@5 / R@10 \\
\hline
ViLBERT-base & 31.86 / 61.12 / 72.80 & - & 58.20 / 84.90 / 91.52 & -  \\
UNITER-base & 66.16 / 88.40 / 92.94 & 80.70 / 95.70 / 98.00 & 72.52 / 92.36 / 96.08 & 85.90 / 97.10 / 98.80  \\
Villa-base & - & - & 74.74 / 92.86 / 95.82 & 86.60 / 97.90 / 99.20  \\
%Ernie-ViL-base & - & - & 74.44 / 92.72 / 95.94 & 86.70 / 97.80 / 99.00   \\
UNIMO-base & 62.44 / 86.16 / 91.68 & 77.40 / 95.10 / 97.80 & 74.66 / 93.40 / 96.08 & 89.70 / 98.40 / 99.10 \\
UNITER-large & 68.74 / 89.20 / 93.86 & 83.60 / 95.70 / 97.70  & 75.56 / 94.08 / 96.76 & 87.30 / 98.00 / 99.20  \\
Villa-large & - & - & 76.26 / 94.24 / 96.84 & 87.90 / 97.50 / 98.80\\
%ERNIE-ViL-large & - & - & 76.70 / 93.58 / 96.44 & 88.10 / 98.00 / 99.20  \\
UNIMO-large & 72.14 / 91.14 / 94.98 & 85.80 / 96.80 / 98.80  & 78.04 / 94.24 / 97.12 & 89.40 / 98.90 / 99.80 \\
ViLT & 51.3 / 79.9 / 81.9 & 69.7 / 91.0 / 96.0 & 62.2 / 87.6 / 93.2 & 83.7 / 97.2 / 98.1 \\
E2E-VLP & - & - & 73.58 / 92.42 / 96.03 & 86.24 / 97.50 / 98.92 \\
SOHO & - & - & 72.5 / 92.7 / 96.1 & 86.5 / 98.1 / 99.3 \\
CLIP & 68.7 / 90.6 / \textbf{95.2} & 88.0 / \textbf{98.7} / \textbf{99.4} & - & - \\
\textbf{Our Baseline} & 65.11 / 87.44 / 92.62 & 78.80 / 94.38 / 97.63 & 78.52 / 94.02 / 96.63 & 91.62 / 98.72 / 99.51\\
\textbf{UNIMO-2} & \textbf{72.70} / \textbf{91.18} / 94.60 & \textbf{88.46} / 96.84 / 98.92 & \textbf{80.14} / \textbf{95.58} / \textbf{97.75} & \textbf{92.01} / \textbf{99.31} / \textbf{99.51} \\
%\hline
%\textbf{UniVLP-large} &  &  &  &  \\
\hline
 \hline
\end{tabular}
\caption{\label{flickr}
Full evaluation results on the Flickr30k retrieval tasks.}
\end{table*}

\section{Experimental Settings}
\label{sec:detail_settings}

\paragraph{Pretraining Datasets} The pre-training datasets consist of text corpus, image collections and image-text pairs. The detail statistics of them are shown in Table \ref{pretrain-dataset}. 

\paragraph{Finetuning Tasks}
\label{sec:finetune}
The multi-modal finetuning tasks include:
\begin{itemize}
	\item \textbf{VQA} requires the model to answer natural language questions by selecting the correct answer from a multi-choice list based on an image. We conduct experiments on the widely-used VQA v2.0 dataset \citep{goyal2017making}, which is built based on the COCO \citep{chen2015microsoft} images. Similar to previous work, both training and validation sets are used for training for the results on both the test-std and test-dev splits.
	\item \textbf{Image Caption} requires the model to generate a natural language description of an image. We report our results on the Microsoft COCO Captions dataset \citep{chen2015microsoft}. Following Karpathy's \citep{karpathy2015deep} split, the dataset contains 113.2k/5k/5k images for train/val/test splits respectively. 
	\item \textbf{Visual Entailment (SNLI-VE)} is evaluated on the SLNI-VE dataset \citep{xie2019visual} which was derived from Flickr30K images and Stanford Natural Language Inference (SNLI) dataset. The task is to determine the logical relationship (i.e., ``Entailment'', ``Neutral'' and ``Contradiction'') between a natural language statement and an image.
	\item \textbf{Image-Text Retrieval} is evaluated on the Flickr30k dataset \citep{young2014image}, which contains two sub-tasks: image retrieval (Flickr30k-IR) and text retrieval (Flickr30k-TR), depending on which modality is used as the retrieved target. We report the top-K retrieval results on the test sets, including R@1, R@5 and R@10.
\end{itemize}

The statistics of the datasets for the above multimodal-tasks are described in Table \ref{finetune-dataset}. 
The hyper-parameters for finetuning all the downstream tasks, including both the single-modal tasks and cross-modal tasks are shown in Table \ref{gen-params} and \ref{multimodal-params}, respectively.
The full evaluation results (including R@1, R@5 and R@10) on Image/Text Retrieval tasks and comparison with other state-of-the-art VLP methods are shown in Table~\ref{flickr}.

\end{document}